\documentclass[12pt]{article}
\usepackage[utf8]{inputenc}
\usepackage[paperheight=29.7cm,paperwidth=21cm,outer=1.5cm,inner=2.5cm,top=2cm,bottom=2cm]{geometry}
\usepackage[table,xcdraw]{xcolor}
\usepackage{authblk}
\usepackage{dirtytalk}
\usepackage[toc,page]{appendix}
\usepackage{graphicx}
\usepackage{xfrac}
\usepackage[hyphens]{url}
\usepackage{adjustbox}
\usepackage{tikz}
\usetikzlibrary{arrows,positioning,shapes.geometric}
\usetikzlibrary{matrix}
\usepackage{pgfkeys}
\usepackage{pgfplots}
\usepgfplotslibrary{groupplots}
\pgfplotsset{compat=newest}
\usepackage[inline]{enumitem}
\usepackage{float}
\usepackage{multirow}
\usepackage{mathtools}
\usepackage{subcaption}
\usepackage{hhline}
\usepackage{cleveref}
\usepackage{soul}

\usepackage{multicol}
\usepackage{float}
\usepackage{fancyhdr}

\newcolumntype{P}[1]{>{\centering\arraybackslash}p{#1}}

 \renewcommand{\paragraph}[1]{
     \textit{#1} 
 }

\fancypagestyle{firstpage}
{
    \fancyhead[C]{\textit{This work has been submitted to the IEEE for possible publication. Copyright may be transferred without notice, after which this version may no longer be accessible.}}    
}

\title{More Real than Real: A Study on \\ Human Visual Perception of Synthetic Faces}
\date{}
\author[1]{Federica Lago}
\author[1]{Cecilia Pasquini}
\author[2]{Rainer Böhme}
\author[3]{Hélène Dumont}
\author[3]{\\Valérie Goffaux}
\author[1]{Giulia Boato}
\affil[1]{Department of Information Engineering and Computer Science, University of Trento}
\affil[2]{Department of Computer Science, University of Innsbruck}
\affil[3]{Institute of Neuroscience, Université Catholique de Louvain}
\begin{document}

\maketitle
\thispagestyle{firstpage}
\textit{Deep fakes} have become popular recently. 
The term refers to doctored media content where one's face is swapped with someone else's face or performs someone else's face movements.
In the last couple of years, numerous video clips, often involving celebrities and politicians, have gone viral on social media platforms.
This has been enabled by easy-to-use apps capable to process user-generated content in real-time.
While early deep fakes were easy to spot, technology has improved and whether or not they can fool a human visual system is still unknown.

Informally, deep fakes can be defined as realistic digital media (images, videos, or audio tracks) depicting untruthful content, obtained either by manipulating pristine material or generated from scratch. The attribute \textit{deep} refers to the use of algorithms based on deep learning, a subfield of modern Artificial Intelligence (AI), which pushed the boundaries for many applications including media data manipulation and generation. 
Besides offering exciting opportunities in several fields (such as entertainment, content production, e-learning, and e-health), these advanced creation technologies are now widely recognized as a pressing threat to the reliability of visual information.\footnote{\url{https://www.theguardian.com/technology/ng-interactive/2019/jun/22/the-rise-of-the-deep fake-and-the-threat-to-democracy}}
This may have severe consequences for the digital identity and reputation of individuals.

Many cases of misuse reported in the past months demonstrate the potential impact of manipulated data on disinformation. 
For example, the last US presidential campaign witnessed the viral diffusion of multiple deceptive visual manipulations: 
amongst others, an altered video of Joe Biden greeting the wrong state in a public speech, as well as retouched pictures of celebrities untruthfully implying their endorsement for Donald Trump.\footnote{\url{https://www.fastcompany.com/90575763/we-have-the-technology-to-fight-manipulated-images-and-videos-its-time-to-use-it}}
Contemporary technologies became so sophisticated that they can build a convincing video of a person based on a single image \cite{tu2021image}, although it is more common to use face reenactment to transfer face movement while preserving the appearance and the identity of the target face \cite{Thies_2016_CVPR}.
Not surprisingly, deep fakes became ubiquitous on the web: the estimated number of deep fakes on the web has doubled every six months since 2018, reaching a total of 85'000 in 2020,\footnote{\url{https://sensity.ai/how-to-detect-a-deep fake-with-sensity/}}
although it is not possible to tell which fraction actually had malicious intents.

Images depicting synthetic faces of non-existing people are easy to produce. They can be created at scale using Generative Adversarial Networks (GANs), that are a variant of deep learning. These images result to be realistic and hard to recognize as synthetic, and thus can be used for scams or to facilitate fraud.\footnote{\url{https://www.ft.com/content/b50d22ec-db98-4891-86da-af34f06d1cb1}} 
Turning back to the 2020 election, a fake report on Joe Biden's son and his connections to China was disseminated by a fabricated digital identity\footnote{\url{https://www.nbcnews.com/tech/security/how-fake-persona-laid-groundwork-hunter-biden-conspiracy-deluge-n1245387}}; the alleged author was a Swiss security analyst, portrayed over the web by a synthetic face presumably created with StyleGAN2 \cite{karras2020analyzing}, the state-of-the-art GAN for still image generation. 
A few months earlier, a 17-year-old student used a GAN-generated face to impersonate and promote the campaign of a fully fictitious congress candidate.\footnote{\url{https://edition.cnn.com/2020/02/28/tech/fake-twitter-candidate-2020/index.html}} 
The hoax was so credible that Twitter activated the ``verified'' icon for the fake account.

Against this backdrop, it is no surprise that different research communities, from multimedia security to computer vision, joined efforts towards the automated detection of AI-generated media. 
A variety of methods have been proposed and refined over the last years \cite{verdoliva2020media} accompanied by the development of benchmark datasets (e.g., FaceForensics++ \cite{rossler2018faceforensics}) and global contests (e.g., Facebook Deep fake Detection Challenge).

While forgeries are almost as old as photography itself, the striking features of the latest advancements in image and video manipulations are the higher accessibility combined with the ever-increasing level of realism. This is particularly true for still images, which today's GANs impressively synthesise through easy-to-use interfaces.\footnote{See for instance \url{https://thispersondoesnotexist.com/}} 
By contrast, the creation of realistic manipulated videos of arbitrary subjects in high resolution still requires significant resources and expertise. 

\begin{figure}[htbp!]
    \centering
    \includegraphics[width=\textwidth]{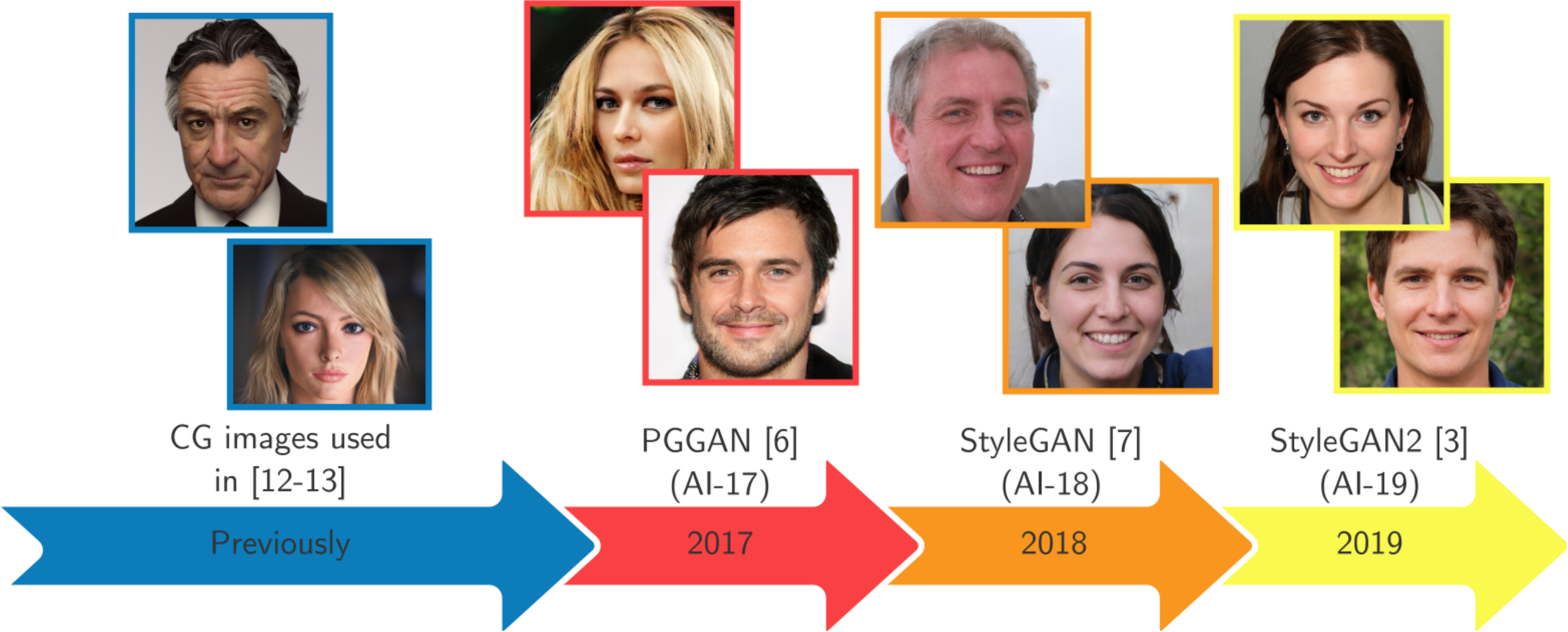}
    \caption{Examples of synthetically generated images over the years. Digital portraits on the left were edited manually, while the following ones are generated through GANs proposed in recent years. We indicate them as AI-17, AI-18, and AI-19 and keep this convention throughout the column.}
    \label{fig:syn-ex}
\end{figure}

In this column, we present a study that measures the human ability to distinguish between real and synthetic face images when confronted with cutting-edge AI-based creation technologies.
\Cref{fig:syn-ex} shows examples of images generated by different GANs: three networks denoted as AI-17 (PGGAN \cite{karras2018progressive}), AI-18 (StyleGAN \cite{karras2019style}) and AI-19 (StyleGAN2 \cite{karras2020analyzing}) are reported. By looking at the manually-edited computer generated portraits on the left, which was the only way to obtain realistic synthetic images until a few years ago, one can appreciate the huge leap forward in terms of human likeness achieved through AIs.
Recent progress is best visible at the level of details.
The latest GAN images no longer contain artifacts and inaccuracies, such as mismatches in eye color, ear shape, teeth rendering, or face symmetry. 
Thus, they evade detection by automated tools relying on exactly these artifacts \cite{verdoliva2020media}.
Outdated algorithms, however, might not be the only ones deceived by last generation data: we ask if human viewers can still tell the difference.

Human perception of manipulated data has been addressed in previous work from several perspectives.
In the field of neuroscience, researchers have investigated how the use of synthetically generated stimuli alters people's ability in face processing tasks, such as face recognition~\cite{crookes2015well}.
In the field of multimedia forensics, researchers have studied to which extent humans perceive face morphing operations, i.e., a specific kind of face manipulation where two faces are blended together to obtain a third hybrid face that carries characteristics of both original subjects. This is particularly relevant for face authentication applications that prevent unauthorized access to places or services, where it is fundamental to study how new morphs affect the ability of humans and algorithms to recognize them as synthetic \cite{makrushin2020simulation}.

Regarding the discrimination of real and synthetic data, the most prominent works are those conducted by Hany Farid and his team, who carried out a series of experiments between 2007 and 2017 on the ability of humans to identify computer generated characters, starting with generic content  \cite{farid2007photorealistic} and then focusing on faces \cite{farid2012perceptual, holmes2016assessing, mader2017identifying}. Their latest studies conclude that humans, on average, are still able to correctly recognize synthetic images. 

While those studies are relatively recent, 
a stringent research question is whether those findings still hold in the light of the striking latest advancements in AI-based synthetic image creation. 
To the best of our knowledge, there is no extensive study on the human perception that involves last-generation high-quality synthetic data.
To address this gap, we have designed and conducted a perceptual experiment where a wide and diverse group of volunteers has been exposed to synthetic face images produced by state-of-the-art GANs  (i.e., AI-17, AI-18, AI-19, in \Cref{fig:syn-ex}).
In the remainder of this column, we report the results of this experiment.
They demonstrate how seriously the human ability to distinguish between synthetic and real should be called into question.

\newpage
\section{Human ratings of real and synthetic faces}

\begin{table}[htbp!]
        \centering
        \begin{tabular}{l|c|ccc|c}
        \hline
        &REAL  & AI-17 & AI-18 & AI-19 & \\ \hline
        Stimuli sample & 150                 & 50    & 50        & 50     & 300   \\
        Participant's sample & 15                  & 5    & 5        & 5  & 30    \\\hline
        \end{tabular}
         \caption{\small Distribution of images belonging to each dataset for the whole stimuli sample and for each randomly generated sequence of images seen by a participant (participant's sample).}
         \label{tab:ds-comp}
\end{table}

The perceptual experiment has been carried out through a dedicated, self-hosted web interface which displays a sequence of stimuli varied between participants. 
The database of stimulus material consisted of 300 images, equally distributed between real and synthetic ones. 
Real images (from now on referred to as REAL) were selected from the FFHQ dataset \cite{karras2019style}; synthetic images were created by the three different GANs, as summarized in \Cref{tab:ds-comp}. 
As mentioned earlier, it is worth observing how the creation technologies suffer from different kinds of visual imperfection (see \Cref{fig:syn-ex}): PGGAN \cite{karras2018progressive} images (AI-17) typically present several face artifacts (e.g., different eye color and size, unnatural hair shape), while StyleGAN \cite{karras2019style} (AI-18) often produces visible blobs, especially in the background; StyleGAN2 \cite{karras2020analyzing} (AI-19) is the most advanced method and overcomes such limitations almost entirely. These three GANs were selected among all the ones proposed recently in the research community because they are among the most used in the literature and they produce highly realistic images that can lead to the previously discussed threats if they are shared online with malicious intents. We wanted a homogeneous set of faces, and therefore the images in each dataset were chosen to balance the gender, to have an age range of about 20-50 years.
We decided to use only Caucasian faces since most datasets are still unbalanced toward this ethnicity.

The experiment was split into five parts:

\begin{itemize}
    \item \textit{Briefing}: participants were informed about the purpose of the study, the expected duration, the target group (18-year-old and above), the voluntary and anonymous nature of the study as well as their rights concerning the protection of personal data. Active consent was sought before proceeding to the next part. 

    \item \textit{Questionnaire}: 
    participants were asked to provide demographic information, to self-assess their ability to recognize faces, and to report their a priori familiarity with deep fakes;
    
    \item \textit{Warm-up phase}: the task of the main experiment was explained and participants had the opportunity to practice with the user interface (see \Cref{fig:screenshot}).
    It displayed a face image sized $10 \times 10$ cm\footnote{Users whose devices were detected to be unfit for this task were not invited to continue the experiment.} 
    ($\approx 4 \times 4$ inches) at the center of the screen for 3 seconds.
    Participants were asked to rate this image by indicating their agreement with the statement ``This image is synthetic" on a 7-point rating scale where the leftmost 1 was labelled ``Completely disagree", (i.e., the image is real) and the rightmost 7 with ``Completely agree'' (i.e., the image is synthetic).
    The midpoint, labelled ``Unsure,'' was selected by default.

    \begin{figure}[htbp!]
        \centering
        \includegraphics[width=0.4\textwidth]{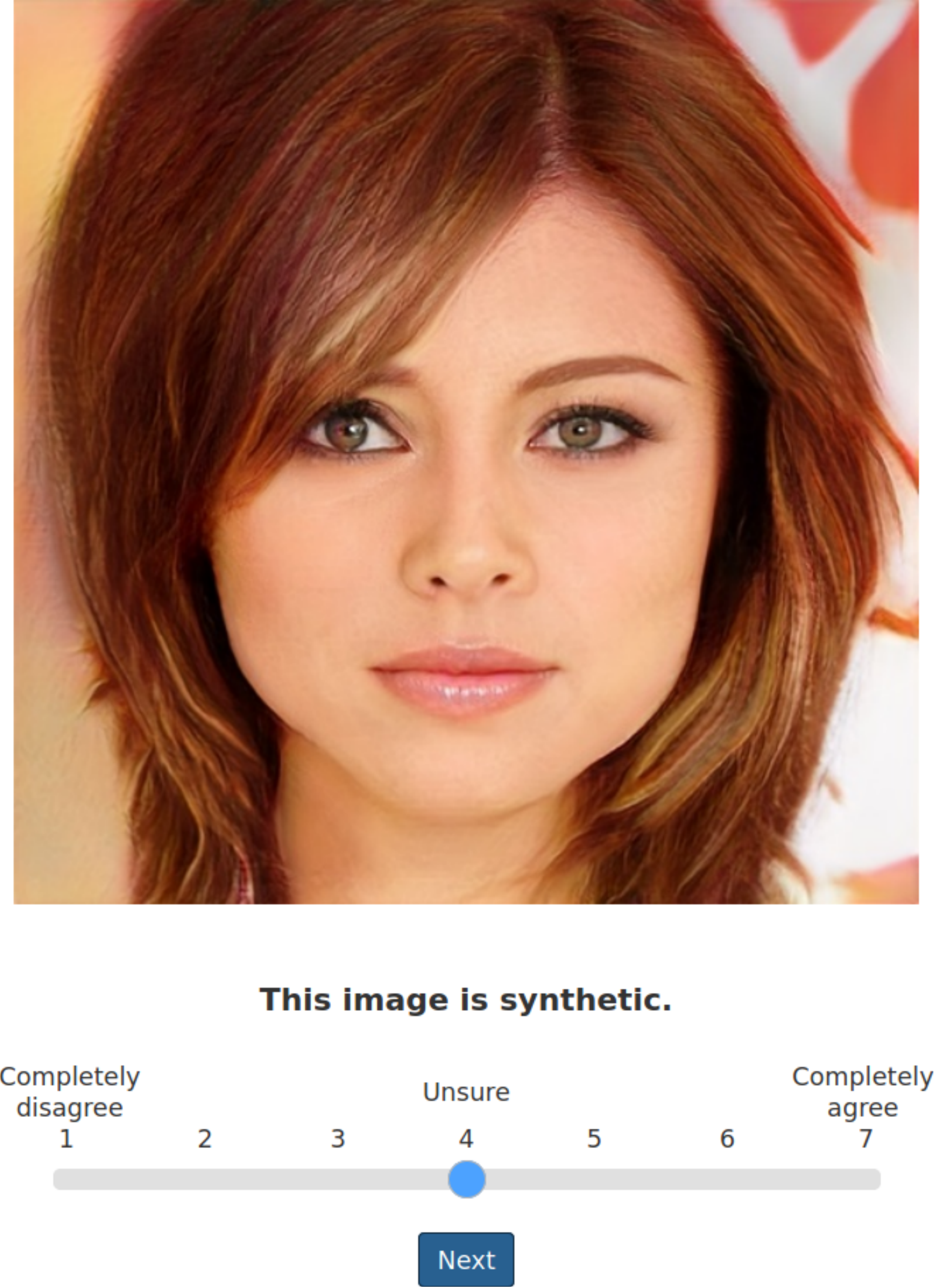}
        \caption{Example screenshot of the user interface for the main task in the English version. The specific face image shown here belongs to AI-17.}
        \label{fig:screenshot}
    \end{figure}

    The response time was unconstrained: participants could select and confirm their rating at any time during or after the 3 seconds in which the image was displayed.
    
   All participants saw the same six images in the warm-up phase. The ratings in this phase were not used in the analysis.
    
    \item \textit{Main experiment}:
    participants sequentially completed a total of 34 task instances.
    A subset of 30 images (the participant's sample) was randomly selected for each participant from the overall collection of stimuli material. The images in each participant's sample were distributed as reported in \Cref{tab:ds-comp} (bottom row). Participants were not informed that images were half real and half synthetic.
    
    The remaining four images served as a control set (two real and two synthetic ones) and were selected to be trivial to rate correctly. The same four images were shown to all participants to check for their compliance with the task.
    
    \item \textit{Exit question and debriefing}:
    finally, participants were invited to describe in an open-ended question their strategy or cues used to rate the images. A response set was stored after confirming consent, followed by a debriefing.
    Participants did not learn about their performance in order to avoid disappointment and to discourage improvement attempts by repeated participation.
    
\end{itemize}

Participants were recruited by extending invitations among professional and personal contacts of all researchers involved (spanning four countries). In order to reach a more diverse sample, the webpage has been made available in four languages: English, French, German, and Italian. Participants completed the experiment remotely on their own devices.

We collected response sets of 630 participants from 38 countries over the field time of 4 months (9$^{th}$ July -- 13$^{th}$ November 2020).
Participants were mostly from Europe (93\%), younger than average (M = 28.15 years, SD = 10.56), and of balanced gender:  45.5\% identified as females, 52.9\% as males, while the remaining preferred to not declare (1.4\%) or identified as non-binary (0.2\%). On average, it took participants 8.43 minutes to complete a response set (median: 7.61).
An estimate of 1000 users have accessed the web interface but did not submit a response set.

In order to limit potential biases, we cleaned the data by discarding participants who exhibited indicators of low interest or distraction.
To this end, we removed 30 (4.76\%) response sets containing outliers in the form of:
\begin{itemize}
    \item skip rate (how many times a participant clicked on the ``Next" button without moving the scale slider);
    \item median response time to rate task instances; or
    \item error rate on the control set (wrong rating on the four control images).
\end{itemize}
We used as cutoffs the 99$^{th}$ percentile of each measure (32.35\% for the skip rate, 7.41 seconds for the median response time, and 3 out of 4 for the error rate on the control set).
A total of 986 (5.48\%) cases where the scale slider was not moved from its default position  were treated as non-response items.

\section{Results}

\begin{figure}[htbp!]
     \centering
     \includegraphics[width=0.9\textwidth]{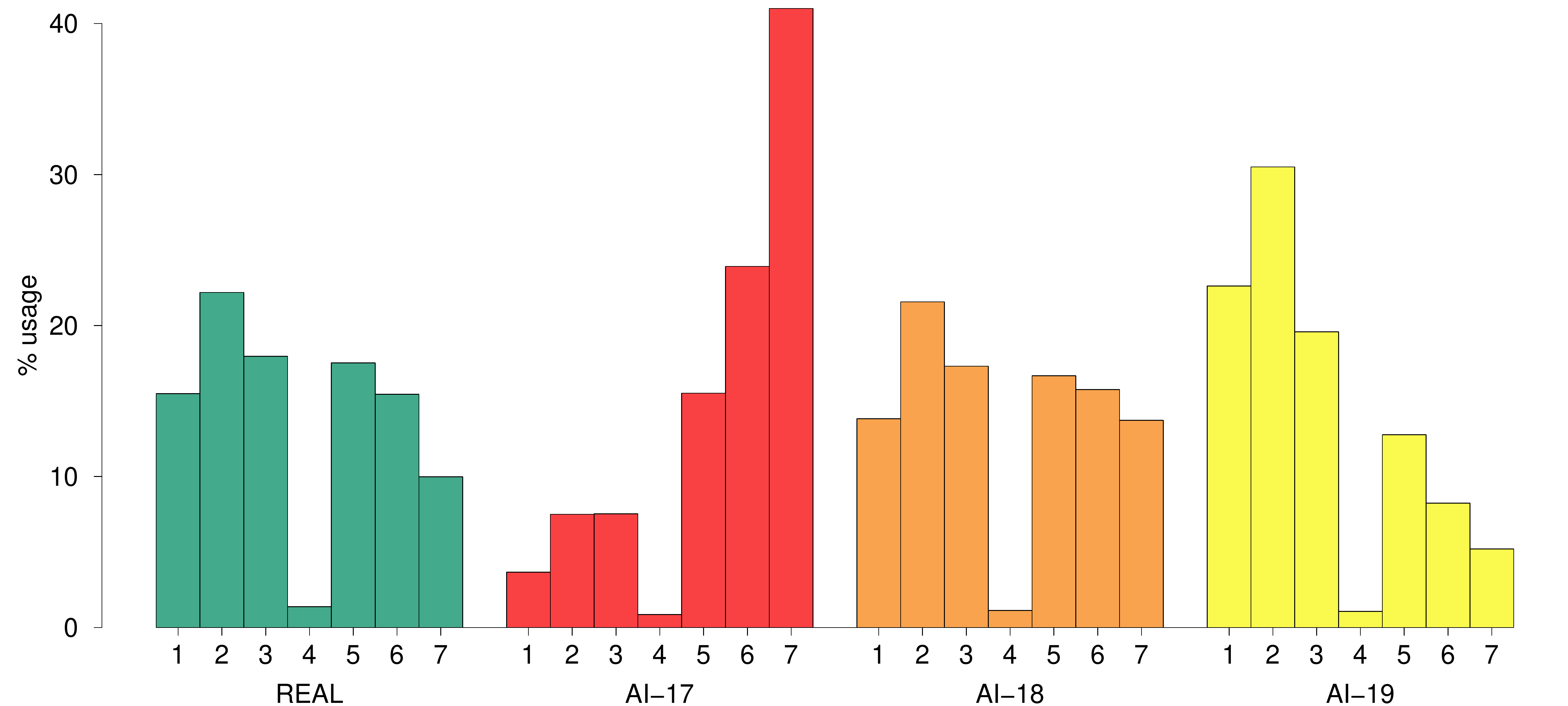}
     \caption{Percentage of scale values (from 1 to 7) used to rate the agreement with the statement ``This image is synthetic". The $N=17014$ ratings from 600 subjects are broken down by datasets.}
    \label{fig:likert}
\end{figure}

\Cref{fig:likert} shows the distribution of the ratings over all participants on images belonging to different datasets. The central value (marked as "Unsure" in the scale) is low because unchanged default values were treated as non-responses, although, in general, participants avoided it.

Observe that the distribution of answers for REAL and AI-18 are very similar, while for AI-17 and AI-19 they are essentially reversed. 
In particular for AI-17, participants were able to correctly recognize the images as synthetic 80\% of the time, of which 41\% with the highest level of agreement with the statement ``This image is synthetic". However, for the successive AIs, the distribution of responses drifts towards the left end of the scale, suggesting that, \textbf{while images generated with earlier AI were still relatively easy to recognize as synthetic, data produced by the newest AIs are increasingly perceived as being real}.
To further investigate the results, we drill down into the dataset using different metrics.

\begin{figure}[htbp!]
    \centering
         \centering
     \begin{subfigure}[b]{0.45\textwidth}
         \centering
         \includegraphics[width=\textwidth]{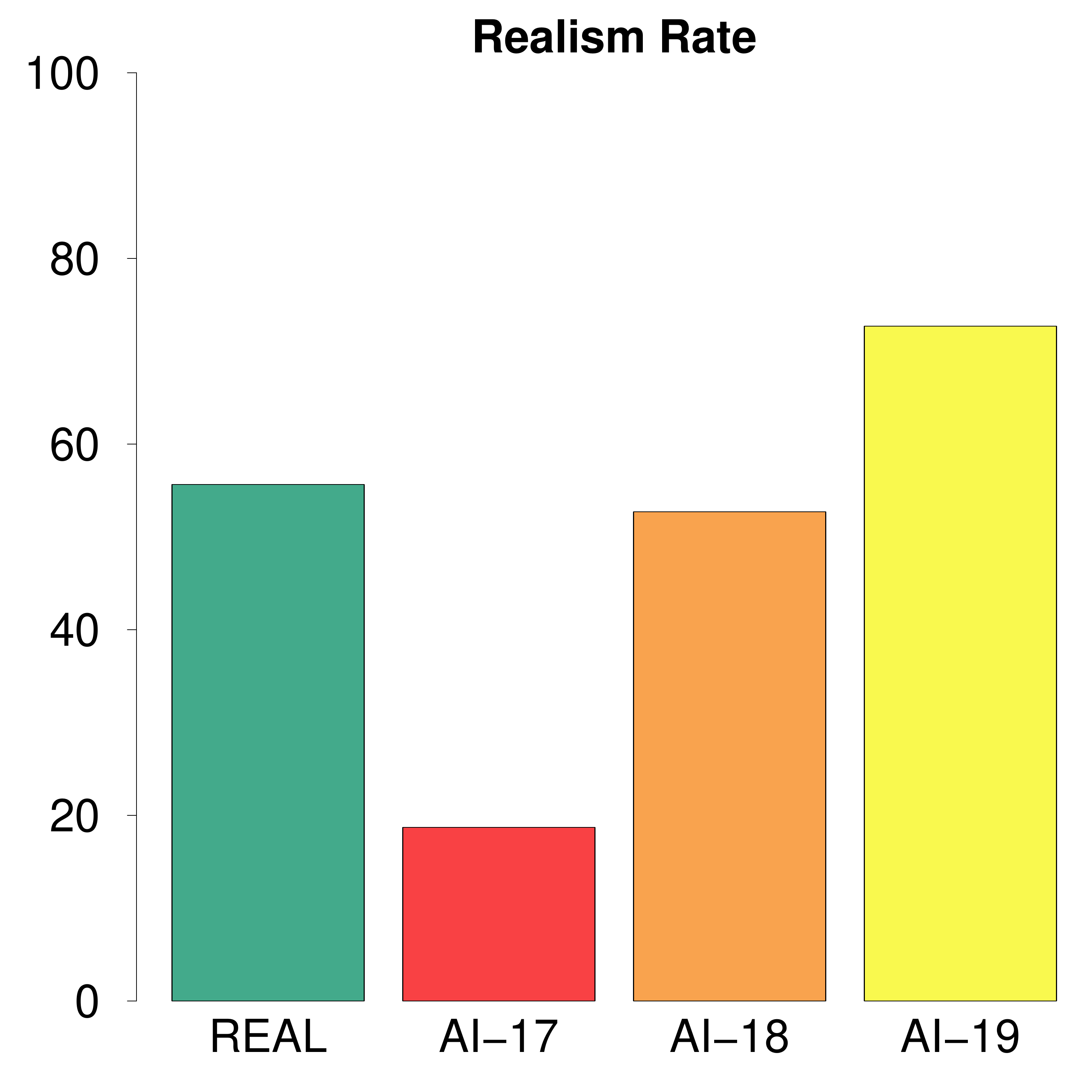}
         \caption{}
         \label{fig:RRA}
     \end{subfigure}
     \begin{subfigure}[b]{0.54\textwidth}
         \centering
         \includegraphics[width=\textwidth]{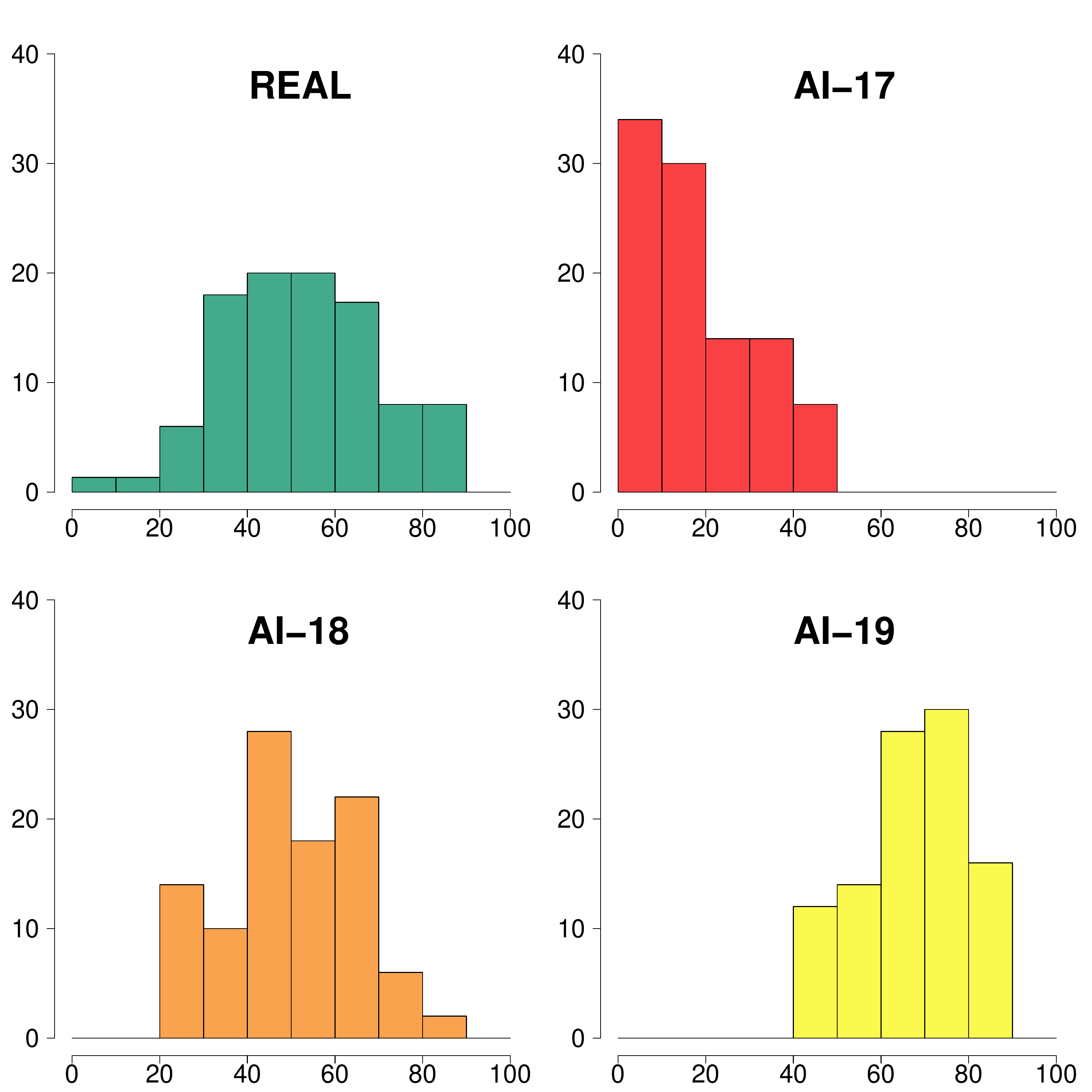}
         \caption{}
         \label{fig:RRB}
     \end{subfigure}
    \caption{(a) shows the \textit{realism rate} per dataset, while (b) shows the distribution of the \textit{realism rates} over the images in each dataset. In (b) the \textit{realism rate} is on the x-axes, while the y-axes indicates the relative frequency of a \textit{realism rate} bin.}
    \label{fig:RR}
\end{figure}

\paragraph{Realism rate.} We aggregate response values in the range from 1 to 3 as a judgement for real and define the \textit{\textbf{realism rate}} as the percentage of images judged as real.
In \Cref{fig:RRA}, the \textit{realism rate} is computed on all the instances of all images belonging to a certain dataset. 
It is striking to observe that the highest \textit{realism rate} (68\%) is achieved by AI-19, while REAL images do not exceed 52\%. The difference is statistically significant (Welch's $t(121)=-6.8$, $p<.001$). In other words, \textbf{synthetic faces generated through StyleGAN2 are judged as real more often than real faces}.

\Cref{fig:RRB} provides a richer visualization as it shows the distribution over $[0,100]$ of \textit{realism rates} of individual images (i.e., computed over multiple task instances involving the same image) of the same dataset. It appears that, \textbf{while some real images were almost never considered to be real (\textit{realism rate} $\approx 0$), all images in AI-19 were judged as real at least 40\% of the time}. Furthermore, skewness values clearly show that, while for REAL and AI-18 the distribution is almost symmetrical (0.016 and 0.149, respectively), the distributions of AI-17 and AI-19 are skewed in opposite directions (0.613 and -0.445, respectively). 

\begin{figure}[htbp!]
    \centering
         \centering
     \begin{subfigure}[b]{0.43\textwidth}
         \centering
         \includegraphics[width=\textwidth]{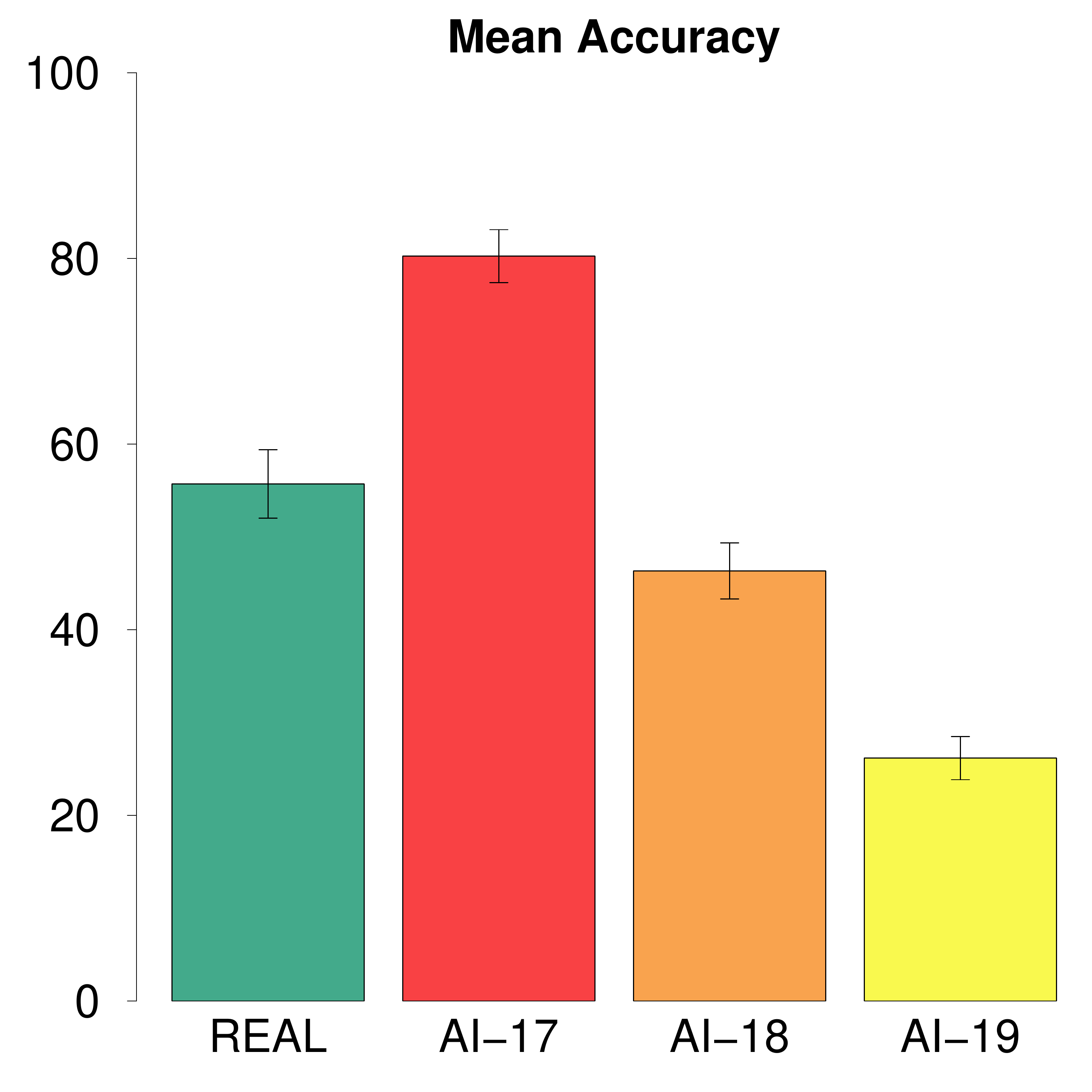}
         \caption{}
         \label{fig:ACC}
     \end{subfigure}
     \begin{subfigure}[b]{0.43\textwidth}
         \centering
         \includegraphics[width=\textwidth]{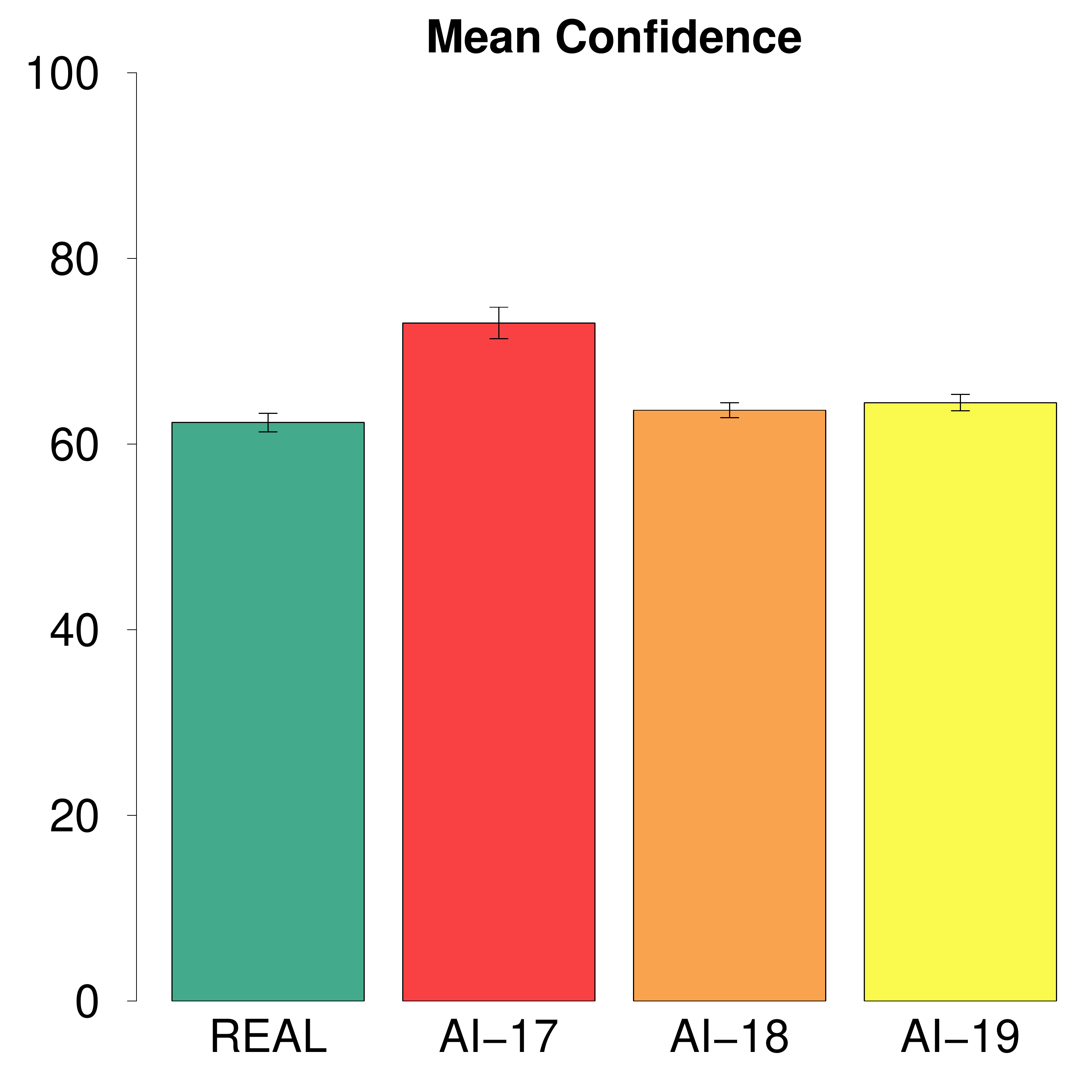}
         \caption{}
         \label{fig:MC}
     \end{subfigure}
    \caption{Comparison of the datasets in terms of (a) mean accuracy and (b) mean confidence. The bars show the average, while the error bars indicate the standard deviation.}
    \label{fig:ACC-MCl}
\end{figure}

%\newpage
\paragraph{Accuracy.} In relation to the \textit{realism rate}, it is interesting to observe the values of the \textit{\textbf{accuracy}} per image, defined as the frequency of correct responses over all task instances with a given image. 
The responses are considered correct if they lie between 1 and 3 for real images or between 5 and 7 for synthetic ones.  
\Cref{fig:ACC} reports the mean values of \textit{accuracy} by dataset, showing that \textbf{the accuracy decreases progressively for newer AIs}. 
The \textit{accuracy} is only slightly above the level of random guessing (56\%) for REAL, while it keeps decreasing for the three synthetic datasets, from 80\% to as little as 26\%.

\medskip
\paragraph{Confidence.} Another indicator is the \textit{\textbf{confidence}}, which is also computed per image. This is obtained by mapping the scale value selected by the participant for a given task instance to the values $[1, \sfrac{2}{3}, \sfrac{1}{3}, 0, \sfrac{1}{3}, \sfrac{2}{3}, 1]$ and by averaging over image. The mapping serves to make the extremes result in the highest confidence and the central value in the lowest confidence.
\Cref{fig:MC} reports the mean values of {the \it confidence} by dataset, showing limited variability both across and within datasets, with a maximum of 73\% for AI-17 and a minimum of 62\% for REAL.

\begin{figure}[htbp!]
    \centering
    \includegraphics[width=0.62\textwidth]{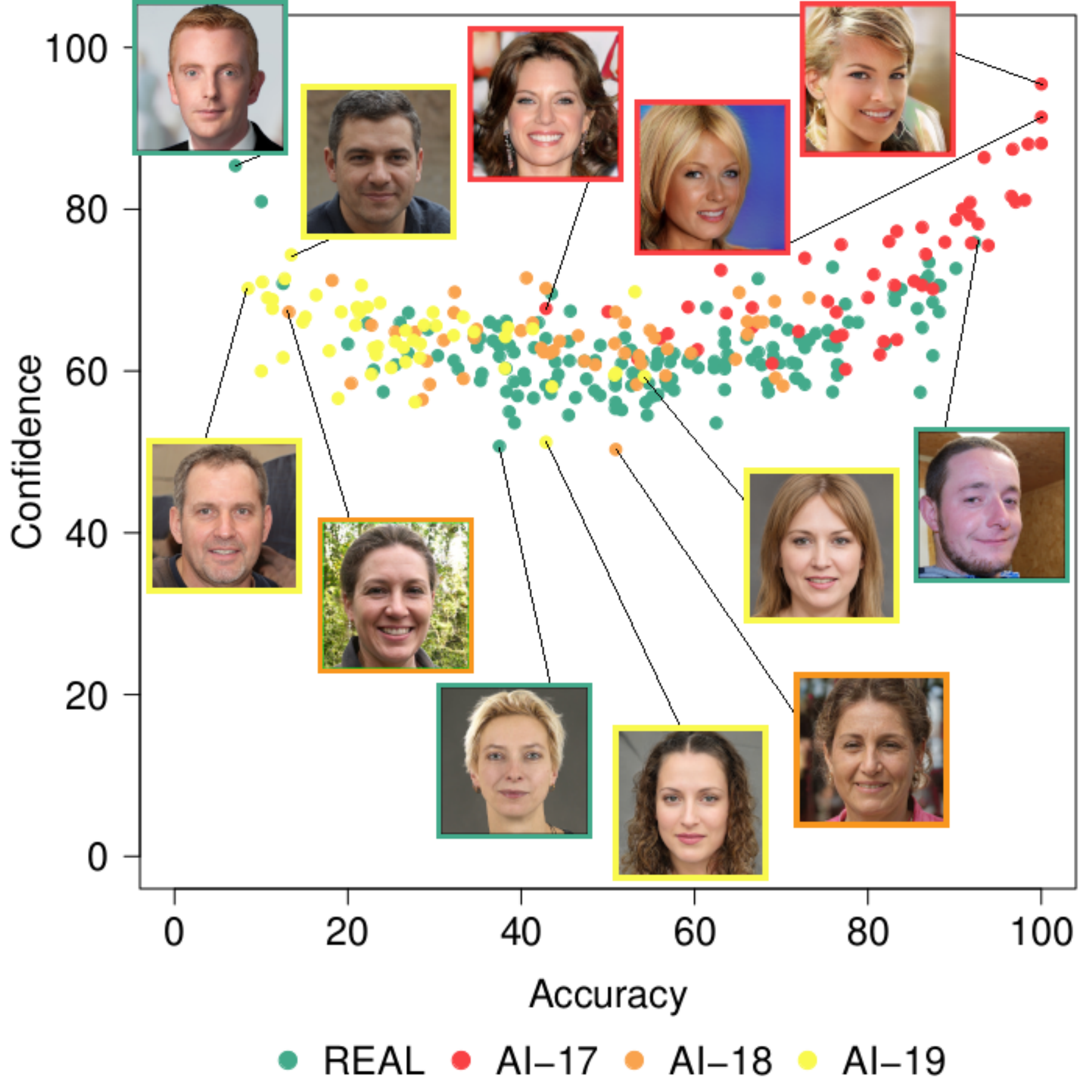}
    \caption{Accuracy (x-axis) and confidence (y-axis) for each image in the stimulus database with selected examples. Color encodes the dataset.}
    \label{fig:im-resp-vs-corr-ex}
\end{figure}

\Cref{fig:im-resp-vs-corr-ex} offers a richer visualization of the same indicators. 
Each dot in the scatter plot represents one image with coordinates defined by its \textit{accuracy} and \textit{confidence}.
Color indicates the dataset.
Selected examples are also reported visually. 
\begin{table}[b!]
\centering
\begin{tabular}{lcccc}
    & \textbf{REAL} & \textbf{AI-17} & \textbf{AI-18} & \textbf{AI-19} \\\hline
$p$  & .006         & \textless .001 & .761          & \textless .001 \\\hline
$\rho$              & .222          & .813           & -.044           & -.462          \\\hline
\end{tabular}
\caption{Spearman correlation results on accuracy and confidence for each dataset.}
\label{tab:spearman}
\end{table}
From the plot, we can observe a ``smile" relationship between \textit{accuracy} and \textit{confidence}. Overall, these two metrics present a weak non-linear correlation (Spearman $\rho = .228$, $p < .001$). By splitting this analysis by dataset (see \Cref{tab:spearman}), it is possible to observe how for REAL and AI-18, that in \Cref{fig:im-resp-vs-corr-ex} present a higher dispersion of the points, \textit{accuracy} and \textit{confidence} are not correlated. On the other hand, for AI-17 and AI-19, whose points are more homogeneously distributed, correlations can be observed. The former has a positive correlation, while the latter presents a negative correlation between \textit{accuracy} and \textit{confidence}. In fact, while images in AI-17 are frequently classified correctly with a high confidence, for AI-19, the more a face is inaccurately perceived as real, the higher the observer's confidence.

%\newpage
\medskip

\paragraph{Indicators of response quality. }
We computed the group median response time for each instance of the main experiment (i.e., from 1 to 34).
We noticed that, \textbf{in the course of the experiment, participants tend to speed up their decisions}: for the first task instance, the median response time was 6.0s, while progressively reducing to 4.6s (median) for the last one. This tendency is confirmed by a significant negative correlation between the task index and the median response time (Spearman $\rho=-.968$, $p < .001$). 
Interestingly, this does not seem to affect the accuracy (Spearman $\rho=-.035$, $p = .843$): \textbf{ participants neither lose accuracy when speeding up, nor do they learn how to correctly classify images in the course of the experiment}.
The latter is quite expected given that participants did not receive any feedback about their performance.
In general, this analysis ensures that our experiment was designed correctly and was not contaminated by a speed/accuracy trade-off.

\medskip
%\newpage
\paragraph{Subject-based analysis.}
We also analyzed the inter-subject variability of the participants by computing the previously defined metrics over the task instances performed by individual subjects.

In general, we found that there is a weak positive correlation between \textit{accuracy} and \textit{confidence} (Spearman $\rho = .201$, $p < .001$), suggesting that participant's confidence is a significant but weak predictor of accuracy. 

This result was further investigated by looking for potential causal factors based on the information collected in the questionnaire.
Participants have been asked to self-evaluate their face recognition skills and their knowledge of the concept of deep fake, on a scale from 1 to 5.\footnote{For the deep fake knowledge, the exact question was \textit{How familiar are you with the concept of deep fakes?}
\begin{enumerate*}
    \item \textit{You did not know they existed}
    \item \textit{You heard about them but never seen any example}
    \item \textit{You have seen examples but have no intuition of the technology behind them}
    \item \textit{You have seen examples and have some knowledge on the technology behind them}
    \item \textit{You have knowledge on the technology behind them and/or have tried to create some yourself.}
\end{enumerate*}} 
Through a linear regression analysis, we found a significant effect of self-reported expertise on the participants' accuracy ($p < .001$): \textbf{participants who are more familiar with deep fakes tend to classify images more accurately than those who are not}. 
Such an effect is however not present for the mean confidence  ($p = .879$), i.e., \textbf{more knowledgeable participants do not seem to provide more confident answers}. Also, self-reported face recognition skills\footnote{For the face recognition skills, the exact question was \textit{How good do you consider yourself to be at remembering faces?}
\begin{enumerate*}
    \item \textit{You tend to forget faces very quickly after seeing them}
    \item \textit{You sometimes mistake people you have met before for strangers}
    \item \textit{You easily recognize faces that you see frequently or occasionally, but you often do not recognize people you have only met briefly before}
    \item \textit{You are better than most people at putting a name to a face}
    \item \textit{You never forget a face.}
\end{enumerate*}} do not significantly influence participants' accuracy nor confidence ($p = .072$ and $p = .100$, respectively).

\medskip

\begin{figure}[t!]
    \centering
     \includegraphics[width=0.75\textwidth]{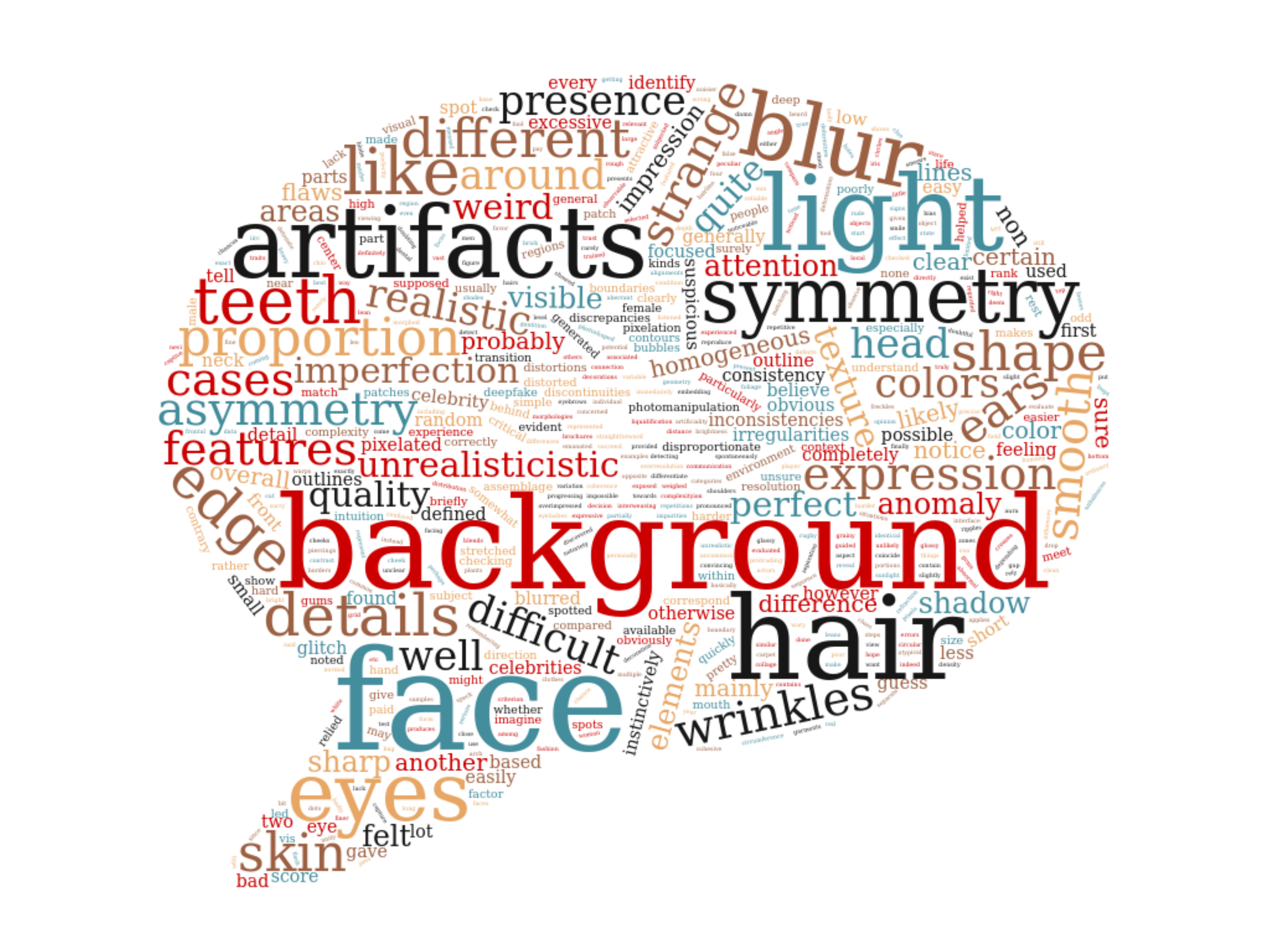}
    \caption{Word-cloud visualization of the reported participants' strategies to recognize synthetic images.}
    \label{fig:wordcloud}
\end{figure}

\paragraph{Participants' strategy.} Finally, we analyzed the participants' free-form answers to the request to describe the strategy used to recognize synthetic images. 
Around 20\% of the participants provided an explanation. 
\Cref{fig:wordcloud} visualizes the answers as a word cloud to provide the reader with a sense of the main keywords that characterize the self-reported strategy of the participants.\footnote{Answers reported in different languages have been translated to English through standard web translation tools.} It can be seen that the background of the image is very important, similarly to the presence of artifacts in general, which also appear as \textit{blur}, \textit{light}, or \textit{anomaly}. Also, specific face elements such as \textit{symmetry}, \textit{teeth}, \textit{hair}, \textit{eyes}, and \textit{wrinkles} are mentioned often.
We also grouped comments according to participants' accuracy ($\ge$ 70\%, between 50\% and 70\%, $\le $50\%) to better analyze them.

Artifacts appear to be a key feature for subjects with accuracy $\geq 0.7$, as they were mentioned in 80\% of the comments. 
These well-performing participants also predominantly reported focusing on the background of the image (67\%) and on the hair (47\%). 
On the contrary, participants with lower performance, and in particular those who performed equal or worse than random guessing, tend to focus less on the background (30\% and 14\%, for those above and below random chance, respectively) and the hair (26\% and 13\%, respectively).
They  concentrate more on the appearance of the face, as roughly 60\% of them mention observing either the general structure of the face (symmetry, proportion, expression) or the realism of facial details, such as eyes, teeth, ears, skin, and wrinkles.
The poorer performance of participants who reported the focus on the face appearance as a strategy stresses once again the striking realism achieved by the most recent AIs.

\medskip

Of course, all results of the subject-based analysis, including the self-reported strategy, must be interpreted with caution as neither the scales were optimized nor the experiment was specifically set up to address the actual image features that determine the realism of a face.
Nevertheless, the self-reported strategies are worth mentioning as they could serve as working hypotheses for future research.
In fact, further investigation is needed to identify relevant aspects of what makes a face look real to human observers, also in relation to the surprisingly high realism rate we observed for StyleGAN2 images with respect to real images.

\section{Comparison to related work}

As mentioned before, the latest works on the ability of people to recognize synthetic images appeared in a 2017 study by Mader, Banks and Farid~\cite{mader2017identifying}. 
The authors used computer generated images edited manually by (human) digital artists as synthetic data.
The real images were collected from different sources.
Among others, a total of 500 participants was tasked with distinguishing between computer generated and real images.
Half of them received a training before task performance was measured (trained), and half of them did not (untrained).

While the different settings preclude a precise quantitative comparison, we can draw qualitative parallels and identify discrepancies on known metrics between our results and the ones reported in \cite{mader2017identifying} for trained and untrained participants.
A first comparison concerns the accuracy per dataset in distinguishing real and synthetic faces: similar values (around 80\%) are found for our participants on AI-17 images and by untrained participants in \cite{mader2017identifying} on CG images. This suggests that the performance of the two populations on the older (although different in nature) kind of manipulations is fairly similar.

\begin{table}[t!]
\centering
\begin{tabular}{l|c|c|c|c|c}
     & \textbf{Untrained \cite{mader2017identifying}} & \textbf{Trained \cite{mader2017identifying}} & \textbf{AI-17}   & \textbf{AI-18}    & \textbf{AI-19} \\\hline
    Sensitivity $d'$   & 1.75 & 2.04 & 1.32            & 0.02            & -0.68                  \\
Bias $\beta$ & 2.36 & 1.45  & 0.7             & 1.05            & 3.39*            \\\hline 
\end{tabular}
\caption{SDT analysis in terms of sensitivity $d'$ and bias $\beta$ reported in \cite{mader2017identifying} -- both with and without training -- and computed on our data for the three synthetic datasets compared to REAL. * indicates that, due to the negative $d'$, $\beta$ was computed on the alternative hypothesis (i.e., the signal is the presence of a real face).}
\label{tab:SDT}
\end{table}

A richer analysis can be obtained using Signal Detection Theory (SDT), which is widely used in experimental psychology to model human detection abilities under uncertainty.
In particular, the sensitivity index ($d'$) quantifies how hard it is to detect the signal (in this case, the presence of a synthetic face), while the bias ($\beta$) measures the extent to which one response is more probable than the other regardless of target presence. In our setting, a value of $\beta > 1$ ($\beta < 1$) indicates that participants are more prone to classify a face as natural (synthetic). The authors of \cite{mader2017identifying} reported that their participants had a clear bias toward classifying an image as real. This bias could only partly be reduced with training (see \Cref{tab:SDT}). 

We decomposed the accuracy computed so far into sensitivity and bias for the three distinct synthetic datasets.
As expected, the analysis shows different results among the three technologies. For AI-17 the sensitivity index is rather high (although always lower than in \cite{mader2017identifying}) and the bias indicates that participants are more prone to consider those faces as synthetic. Sensitivity and bias for AI-18 indicate a performance close to random guessing. Finally, the negative sensitivity registered for AI-19, along with the extremely high bias registered for the alternative hypothesis, show to what extent participants are misled into thinking that those faces are real.

\section{Discussion and future work}

This study provides first quantitative evidence on how the quality and realism of face images generated with cutting-edge AIs makes it hard for human viewers to recognize them as synthetic. This trend is rather prominent as, within just two years, we moved from synthetic images that were reliably detected (82\% accuracy for AI-17) to synthetic images whose realism rate even surpasses real images (68\% for AI-19 versus 52\% for REAL).
In other words, our participants questioned the authenticity of recent GAN-generated faces less than they did for images of real faces!

The fact that recent AI has jumped this bar so impressively poses a string of follow-up questions.
One interesting direction would be to study how these AI-based creation technologies position themselves with regards to the \textit{uncanny valley} effect.
This well-known phenomenon in robotics and computer graphics predicts that people's response to human-like avatars quickly shifts from empathy to revulsion as the avatars approach, but do not fully attain, a human appearance~\cite{mori2012}.
In fact, our results suggest that the recent developments have pushed the synthetic face generation technology away from the uncanny valley, calling for a rise in awareness among users on the existence, threats and opportunities of deep fakes.

On the one hand, there is the need for new automated methods capable to detect synthetic faces in applications where authenticity matters, such as authentication services or the media industry, but also on consumer devices where content is viewed to reduce the dissemination and harmfulness of fake news. However, automatic detectors might not ever work sustainably, given the way GANs work: every improved detector can be used as a discriminator and make the next generation GAN evade it altogether.

On the other hand, deep fakes could provide a valuable resource for fields such as cognitive neuroscience, that currently are often limited by the scarcity of extensive sets of controlled face images. The perception of faces by humans is a core research topic in the vast field of the cognitive neuroscience, as a substantial portion of the human visual system is dedicated to perceiving face signals; the cortical territory devoted to the visual perception of other visual categories (e.g., natural scene images) is much less extensive. Human newborns, since the very first minutes of life, preferentially turn their gaze towards faces. In fact, faces emit rich and complex signals that inform on others’ intentions, trustworthiness, center of attention, identity, age, emotional expressions, physical health, all cues that are important for an individual's social life quality.

The current results open exciting avenues for the generation of extensive face stimulus sets for human face perception research. As recent research suggests \cite{vanrullen2019reconstructing}, the investigation and manipulation of the most recent deep fakes latent spaces promises to further deepen our understanding of how the human brain perceives faces.

\section*{Ethics}
The experimental protocol adhered to the Declaration of Helsinki and was approved by the local ethical committee (Psychological Sciences Research Institute, UCLouvain). [no official reference number attributed by this commission]

\section*{Acknowledgement}
This work was supported by the project PREMIER (PRE-serving Media trustworthiness in the artificial Intelligence ERa), funded by the Italian Ministry of Education, University, and Research (MIUR) within the PRIN 2017 program.\\
V.G. is a research associate of the National Fund for Scientific Research (F.R.S.-FNRS).\\
This study has received funding from the FWO and F.R.S.-FNRS under the Excellence of Science (EOS) programme (HUMVISCAT-30991544) awarded to V.G..

%\clearpage
%\newpage
\bibliographystyle{unsrt}
\bibliography{refeences}

\begin{thebibliography}{10}

\bibitem{tu2021image}
Xiaoguang Tu, Yingtian Zou, Jian Zhao, Wenjie Ai, Jian Dong, Yuan Yao, Zhikang
  Wang, Guodong Guo, Zhifeng Li, Wei Liu, et~al.
\newblock {Image-to-Video Generation via 3D Facial Dynamics}.
\newblock {\em IEEE Transactions on Circuits and Systems for Video Technology},
  2021.

\bibitem{Thies_2016_CVPR}
Justus Thies, Michael Zollhofer, Marc Stamminger, Christian Theobalt, and
  Matthias Niessner.
\newblock {Face2Face: Real-Time Face Capture and Reenactment of RGB Videos}.
\newblock In {\em Proceedings of the IEEE Conference on Computer Vision and
  Pattern Recognition (CVPR)}, 2016.

\bibitem{karras2020analyzing}
Tero Karras, Samuli Laine, Miika Aittala, Janne Hellsten, Jaakko Lehtinen, and
  Timo Aila.
\newblock Analyzing and improving the image quality of stylegan.
\newblock In {\em Proceedings of the IEEE/CVF Conference on Computer Vision and
  Pattern Recognition}, pages 8110--8119, 2020.

\bibitem{verdoliva2020media}
Luisa Verdoliva.
\newblock Media forensics and deepfakes: An overview.
\newblock {\em IEEE Journal of Selected Topics in Signal Processing},
  14(5):910--932, 2020.

\bibitem{rossler2018faceforensics}
Andreas R{\"o}ssler, Davide Cozzolino, Luisa Verdoliva, Christian Riess, Justus
  Thies, and Matthias Nie{\ss}ner.
\newblock Faceforensics: A large-scale video dataset for forgery detection in
  human faces.
\newblock {\em arXiv preprint arXiv:1803.09179}, 2018.

\bibitem{karras2018progressive}
Tero Karras, Timo Aila, Samuli Laine, and Jaakko Lehtinen.
\newblock Progressive growing of gans for improved quality, stability, and
  variation.
\newblock In {\em International Conference on Learning Representations}, 2018.

\bibitem{karras2019style}
Tero Karras, Samuli Laine, and Timo Aila.
\newblock A style-based generator architecture for generative adversarial
  networks.
\newblock In {\em Proceedings of the IEEE Conference on Computer Vision and
  Pattern Recognition}, 2019.

\bibitem{crookes2015well}
Kate Crookes, Louise Ewing, Ju-dith Gildenhuys, Nadine Kloth, William~G
  Hayward, Matt Oxner, Stephen Pond, and Gillian Rhodes.
\newblock How well do computer-generated faces tap face expertise?
\newblock {\em PloS one}, 10(11):e0141353, 2015.

\bibitem{makrushin2020simulation}
Andrey Makrushin, Dennis Siegel, and Jana Dittmann.
\newblock Simulation of border control in an ongoing web-based experiment for
  estimating morphing detection performance of humans.
\newblock In {\em Proceedings of the 2020 ACM Workshop on Information Hiding
  and Multimedia Security}, pages 91--96, 2020.

\bibitem{farid2007photorealistic}
Hany Farid and Mary Bravo.
\newblock Photorealistic rendering: How realistic is it?
\newblock {\em Journal of Vision}, 7(9):766--766, 2007.

\bibitem{farid2012perceptual}
Hany Farid and Mary~J Bravo.
\newblock Perceptual discrimination of computer generated and photographic
  faces.
\newblock {\em Digital Investigation}, 8(3-4):226--235, 2012.

\bibitem{holmes2016assessing}
Olivia Holmes, Martin~S Banks, and Hany Farid.
\newblock Assessing and improving the identification of computer-generated
  portraits.
\newblock {\em ACM Transactions on Applied Perception (TAP)}, 13(2):1--12,
  2016.

\bibitem{mader2017identifying}
Brandon Mader, Martin~S Banks, and Hany Farid.
\newblock Identifying computer-generated portraits: The importance of training
  and incentives.
\newblock {\em Perception}, 46(9):1062--1076, 2017.

\bibitem{mori2012}
Masahiro Mori, Karl~F MacDorman, and Norri Kageki.
\newblock The uncanny valley [from the field].
\newblock {\em IEEE Robotics \& Automation Magazine}, 19(2):98--100, 2012.

\bibitem{vanrullen2019reconstructing}
Rufin VanRullen and Leila Reddy.
\newblock Reconstructing faces from f{MRI} patterns using deep generative
  neural networks.
\newblock {\em Communications {B}iology}, 2(1):1--10, 2019.

\end{thebibliography}

%% CREATES 3 COLUMNS
%\end{multicols}

%\newpage
%\input{auth}

%\newpage
%\begin{appendices}
%\input{app}
%\end{appendices}
\end{document}